\documentclass[10pt,twocolumn,letterpaper]{article}

\usepackage{iccv}              

\usepackage{times}
\usepackage{epsfig}
\usepackage{graphicx}
\usepackage{amsmath}
\usepackage{amssymb}

\usepackage{amsfonts}
\usepackage{array}
\usepackage{textcomp}
\usepackage{stfloats}
\usepackage{url}
\usepackage{verbatim}
\usepackage[numbers]{natbib}

\usepackage{soul} 
\usepackage{lipsum}
\usepackage{xcolor}
\definecolor{lightblue}{rgb}{0.8,0.9,1}
\definecolor{lightgreen}{rgb}{0.9,1,0.9}
\definecolor{lightred}{rgb}{1,0.5,0.5}
\definecolor{gray}{rgb}{0.7,0.7,0.7}

\usepackage{multirow}

\usepackage{amsthm}

\usepackage[ruled]{algorithm}
\usepackage{algpseudocode}

\usepackage{./rpm_packages/rpm_SIunits}
\usepackage{./rpm_packages/rpm_acronyms}
\usepackage{./rpm_packages/rpm_math}
\usepackage{./rpm_packages/rpm_misc}

\usepackage{kotex}

\definecolor{iccvblue}{rgb}{0.21,0.49,0.74}
\usepackage[pagebackref,breaklinks,colorlinks,allcolors=iccvblue]{hyperref}

\def\paperID{5260} 
\def\confName{ICCV}
\def\confYear{2025}

\title{THE-Pose: Topological Prior with Hybrid Graph Fusion\\for Estimating Category-Level 6D Object Pose}

\author{Eunho Lee\\
Seoul National University\\
{\tt\small eunho1124@snu.ac.kr}
\and
Chaehyeon Song\\
Seoul National University\\
{\tt\small chaehyeon@snu.ac.kr}
\and
Seunghoon Jeong\\
Seoul National University\\
{\tt\small shoon0602@snu.ac.kr}
\and
Ayoung Kim\\
Seoul National University\\
{\tt\small ayoungk@snu.ac.kr}
}

\begin{document}
\maketitle
\begin{abstract}

Category-level object pose estimation requires both global context and local structure to ensure robustness against intra-class variations. However, 3D graph convolution (3D-GC) methods only focus on local geometry and depth information, making them vulnerable to complex objects and visual ambiguities. 
To address this, we present THE-Pose, a novel category-level 6D pose estimation framework that leverages a topological prior via surface embedding and hybrid graph fusion. Specifically, we extract consistent and invariant topological features from the image domain, effectively overcoming the limitations inherent in existing 3D-GC based methods. 
Our Hybrid Graph Fusion (HGF) module adaptively integrates the topological features with point-cloud features, seamlessly bridging 2D image context and 3D geometric structure. These fused features ensure stability for unseen or complicated objects, even under significant occlusions.
Extensive experiments on the REAL275 dataset show that THE-Pose achieves a 35.8\% improvement over the 3D-GC baseline (HS-Pose) and surpasses the previous state-of-the-art by 7.2\% across all key metrics. The code is avaialbe on \url{https://github.com/EHxxx/THE-Pose}
\end{abstract}


\section{Introduction}
\label{sec:intro}



\noindent
6D object pose estimation aims to obtain 3D location and rotation of the objects in a single RGB-D image, which is crucial for robotics manipulation~\cite{MN1, MN2}, augmented reality~\cite{AR1, AR2, AR3}, and scene understanding~\cite{SU1, SU2}. Early work focuses on instance-level pose estimation, where each object has a dedicated 3D CAD model~\cite{Posecnn, Densefusion, SurfEmb, Bb8, PVNet}. Although these approaches perform well for known objects, they rely heavily on instance-specific 3D models and fail to generalize to novel objects in practical scenarios, where collecting CAD models is infeasible. In contrast, category-level 6D pose estimation~\cite{NOCS, SPD, FS-Net, HS-Pose, AG-Pose, SecondPose} attempt to predict rotation, translation, and size for unseen objects in a given category, eliminating the need for individual CAD models. The primary challenge in this field is the catastrophic intra-class variations, including diverse shape and color differences within each category.

\begin{figure}[!t]
  \centering
   \includegraphics[trim=60 40 55 120, clip, width=0.9\columnwidth ]{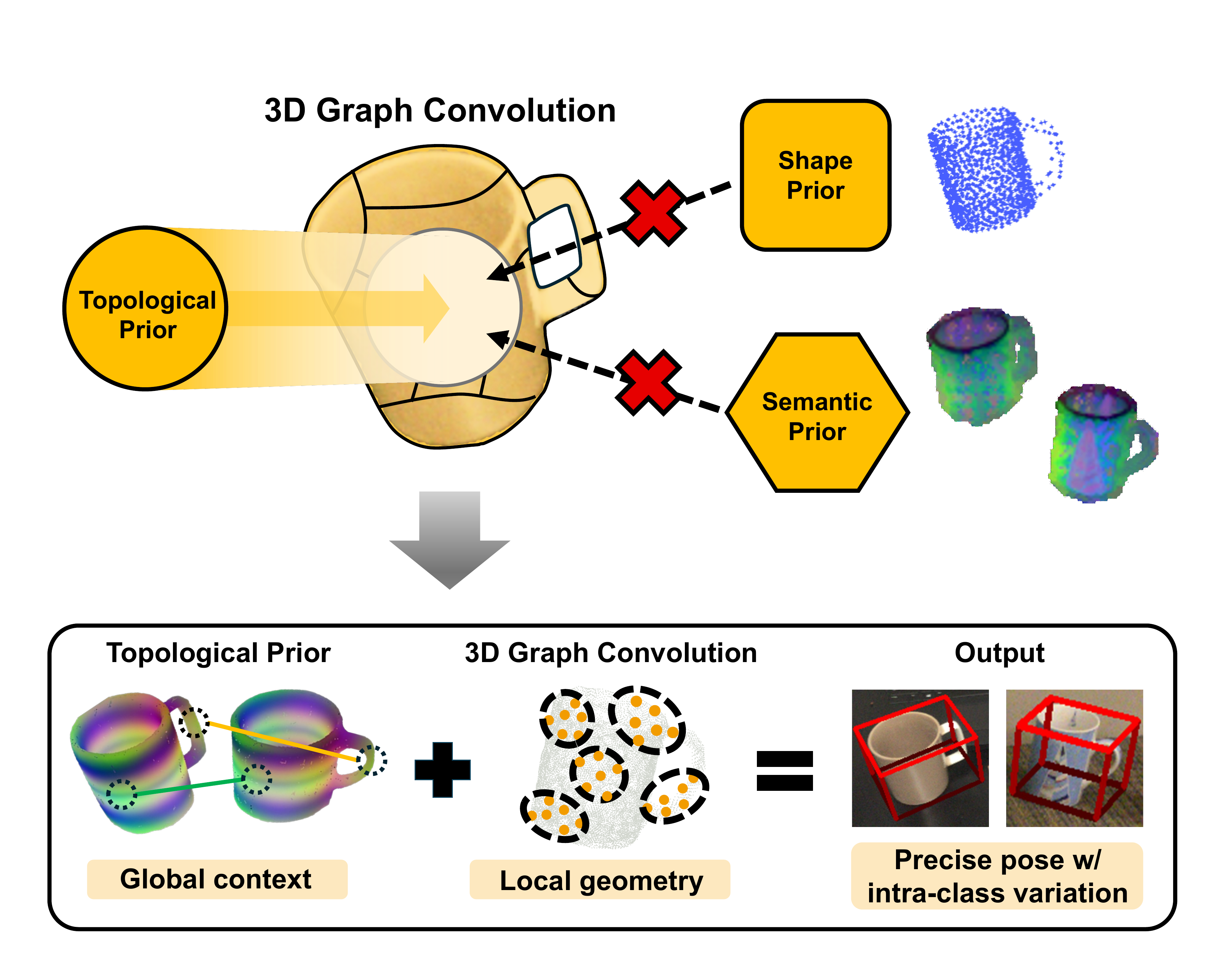}
  \caption{\textbf{Overview}. We overcome the limitations of 3D graph convolution methods by proposing the topological prior. Since other types of priors, such as shape prior and semantic prior, contain inappropriate and unnecessary information for category-level 6D pose estimation, they often degrade overall performance. However, as our prior focuses on global geometry and topology, we successfully combine it with the 3D-GC module, resulting in astonishing performance and overcoming the limitations of 3D-GC.}
  \label{fig:overview}
  \vspace{-3mm}
\end{figure}

To address this issue, a concept of categorical prior is introduced, which features common information for each category.  
By utilizing RGB or RGB-D features, most works~\cite{SPD, RBP-Pose, Self-DPDN} succeed to obtain the correspondences between the observed object and the shape prior derived from a mean shape of each category. However, the averaging attribute leads to numerous incorrect correspondences under significant shape variation~\cite{AG-Pose}. Recently, SecondPose~\cite{SecondPose} attempted to overcome these limitations by employing semantic priors derived from a DINOv2~\cite{DINOv2}. Unfortunately, as the vision foundation model is developed for general tasks, it includes unnecessary visual information, causing inconsistent performance under significant texture variation.

On the other hand, 3D graph convolution (3D-GC) based methods~\cite{FS-Net, vlach2016we} argue that relying on RGB features can degrade performance due to large color diversity. Hence, depth-only 3D-GC approaches~\cite{3D-GC, FS-Net, GPV-Pose, HS-Pose} directly regress 6D poses from point clouds and show local geometric awareness and better generalization ability to unseen objects. However, their focus is limited to local structures without considering the global context, which makes them vulnerable to outliers and less suitable for categories that have complex shapes. Moreover, the scale- and translation-invariant nature of 3D-GC hinders accurate estimation of object size and translation.

Inspired by the above discussions, we address the limitations of 3D-GC methods by introducing a new type of prior called the \textbf{Topological Prior} as \hyperref[fig:overview]{{\underline{\textcolor{blue}{Figure } \textcolor{red}{1}}}}. Conventionally, the image domain was considered unhelpful for existing 3D-GC based approaches since previous priors focused on subsidiary color information. We conclude this confusion by developing the topological prior, which provides only essential geometric information as the categorical prior from RGB image. We expand the surface embedding~\cite{SurfEmb}, which is designed to learn 2D–3D correspondences for category-level. 
With this prior, we propose \textbf{THE-Pose}: \textbf{T}opological Prior with \textbf{H}ybrid Graph Fusion for \textbf{E}stimating Category-Level 6D \textbf{Pose}. It successfully integrates the topological prior and graph convolution through two sub-modules.
First, the \ac{TGC} module builds topological prior to obtain global context for overcoming 3D-GC's limitations and provide vital information to the graph-convolution stage called \ac{HGF} module. In the HGF, topological features and point cloud features are adaptively fused via a hybrid receptive field and then aggregated into a hierarchical representation.

As a result, our method improves overall performance and robustness, which substantiates the potential of integrating the image-driven topological prior and 3D-GC. Our main contributions are summarized as follows:

\begin{enumerate}
    \item We define a novel prior called \textbf{Topological prior}. This prior satisfies crucial geometric constraints for category-level 6D pose estimation and fills the missing piece in 3D graph convolution.
    
    \item We propose \textbf{THE-Pose}, which effectively captures both local and global geometric structures by adaptively fusing topological and point cloud features. Our method effectively handles intra-class variations and is robust to complex shapes and visual ambiguities.

    
    
    \item Extensive experiments on the CAMERA25, REAL275 datasets demonstrate the potential of combining topological prior and 3D-GC while achieving a new state-of-the-art performance.
    
\end{enumerate}

\section{Related works}
\label{sec:related}

\begin{figure*}[t]
  \centering
  \includegraphics[width=1\textwidth]{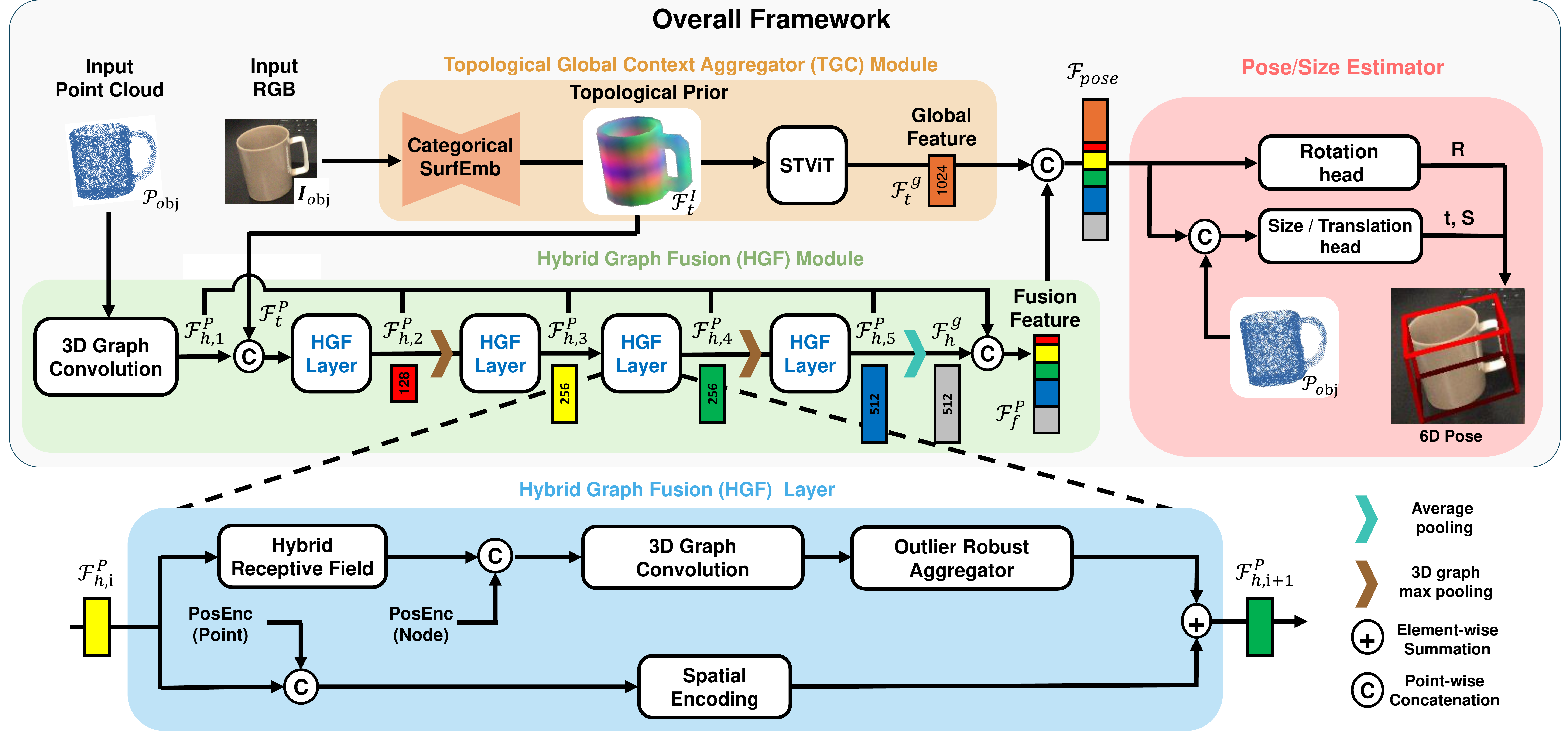}
    \caption{\textbf{Overvall Framework of the proposed THE-Pose.} The TGC module extracts topological prior features \(\mathcal{F}^I_t\) and global features \(\mathcal{F}^g_t\) using the categorical SurfEmb model. Simultaneously, 3D graph convolution (3D-GC) produces local point-cloud features \(\mathcal{F}^P_{h,1}\). In the HGF module, these feature are fused through a hybrid receptive field and 3D-GC layers, generating hierarchical fused features \(\mathcal{F}^g_f\). Finally, we concatenate \(\mathcal{F}^g_t\) and \(\mathcal{F}^P_f\) to form the pose feature \(\mathcal{F}_{\text{pose}}\) for 6D pose estimation.}

  \label{fig:main-pipeline}
\end{figure*}

\subsection{Prior Based Methods}
Category-level 6D pose estimation aims to predict rotation, translation, and size for unseen objects within a given category. A key challenge is the significant intra-class variation in shape and texture. To address this, \citet{NOCS} introduced Normalized Object Coordinate Space (NOCS) as a canonical representation. This approach involves mapping the observed point cloud to the NOCS system, followed by pose recovery via the Umeyama algorithm~\cite{umeyama1991least}. SPD~\cite{SPD} subsequently improved upon this by deforming a categorical shape prior into the NOCS representation, aligning partial observations to establish dense correspondences. Motivated by SPD, several subsequent approaches~\cite{SPD, Cascaded, Sgpa, 6D-vit, ACR-Pose, Catformer} explored various strategies for enhancing shape prior deformation and correspondence estimation, leading to performance improvements. SecondPose~\cite{SecondPose} recently demonstrated the effectiveness of semantic priors from foundation models such as DINOv2~\cite{DINOv2} in handling intra-class variations. However, their reliance on 2D features limits their ability to be susceptible to texture variations and capture detailed 3D geometric structures. In contrast, our approach employs a novel topological prior that effectively preserves 3D geometric consistency, addressing limitations of existing methods and robustly handling intra-class variations.

\subsection{3D Graph Convolution Based Methods}

Recent advances in 3D Graph Convolution (3D-GC)~\cite{3D-GC} have significantly improved category-level pose estimation. Depth-only 3D-GC methods~\cite{FS-Net, GPV-Pose, HS-Pose} argue that texture variations are more problematic than shape variations, thus directly regressing object poses from point clouds without using RGB features. FS-Net~\cite{FS-Net} estimates rotation, translation, and size through direct regression, while GPV-Pose~\cite{GPV-Pose} further incorporates geometric priors through a symmetry-aware reconstruction network and point-wise voting. However, these approaches are inherently limited by the local scope of 3D-GC, lacking sufficient global context and accurate scale and translation estimation. To address this, HS-Pose~\cite{HS-Pose} introduced a hybrid feature extraction layer (HS-layer) that perceives both local and global geometry, encodes translation and scale information, and remains robust to outliers. By replacing the 3D-GC layers in GPV-Pose~\cite{GPV-Pose} with the HS-layer, HS-Pose effectively handles complex object shapes and achieves real-time performance. Nonetheless, there remains substantial room for improvement in their fundamental limitations.
 To overcome these issues, we propose integrating RGB-based topological priors within the 3D-GC framework, effectively addressing these limitations and enhancing robustness.

\section{Methodology}
\label{sec:methods}

\subsection{Overall Framework} 
\label{subsec:overview}

Given an RGB-D image, we first use a segmentation model, such as Mask R-CNN \cite{MaskRcnn}, to obtain each object's segmentation mask and category label. For each segmented object, we extract a cropped RGB image $\mathbf{I}_{\text{obj}} \in \mathbb{R}^{H \times W \times 3}$ and generate a point cloud $\mathcal{P}_{\text{obj}} \in \mathbb{R}^{N \times 3}$ by back-projecting the depth image using camera intrinsics and applying a downsampling process. Using these inputs, THE-Pose predicts the full 6D pose of the object, including 3D rotation $\mathbf{R} \in SO(3)$, 3D translation $\mathbf{t} \in \mathbb{R}^3$, and 3D size $\mathbf{s} \in \mathbb{R}^3$.

The framework of THE-Pose is shown in \hyperref[fig:main-pipeline]{{\underline{\textcolor{blue}{Figure} \textcolor{red}{2}}}}. The method consists of three main components:
i) \textbf{Topological Global Context Aggregator} (\hyperref[subsec:TGC]{{\underline{\textcolor{blue}{Sec.} \textcolor{red}{3.2}}}}), ii) \textbf{Hybrid Graph Fusion module} (\hyperref[subsec:HGF]{{\underline{\textcolor{blue}{Sec.} \textcolor{red}{3.3}}}}), iii) \textbf{Pose\&Size Estimator} (\hyperref[subsec:PSE]{{\underline{\textcolor{blue}{Sec.} \textcolor{red}{3.4}}}}).
Details about each component are provided in the following sections.

\begin{figure}[!t]
  \centering
  \includegraphics[width=\linewidth]{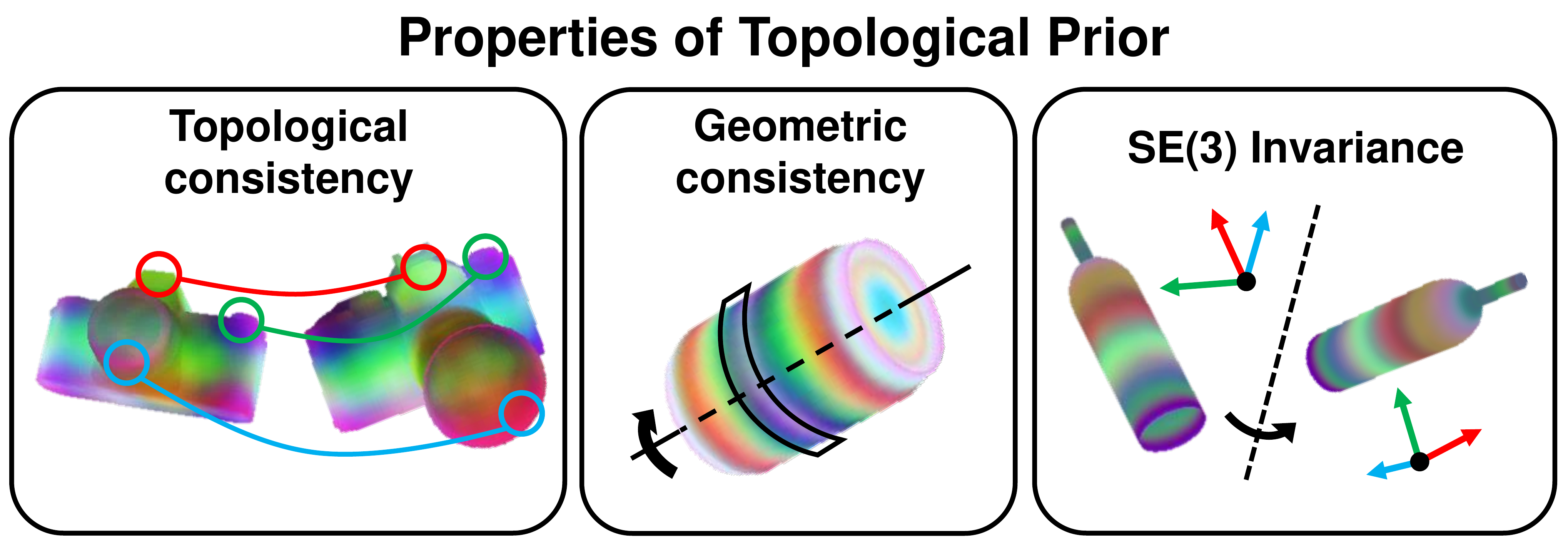}
  \caption{\textbf{Topological Prior}. Our prior satisfies three properties. Topologically equivalent parts within the category are encoded with consistent representations (Topological consistency). Symmetric surfaces are represented by identical features (Geometric consistency). The prior remains unaffected by rotation and translation ($SE(3)$ invariance).}
  \label{fig:topological_prior}
\end{figure}

\subsection{Topological Prior and Global Context}
\label{subsec:TGC}
Irrelevant information in priors often limits the overall performance. To seamlessly combine with and enhance the 3D graph convolution (3D-GC), the prior should possess only primary geometric information of the object. We have discussed the optimal form of the prior and finally developed the \textbf{Topological Prior}, which has three properties as \hyperref[fig:topological_prior]{{\underline{\textcolor{blue}{Figure } \textcolor{red}{3}}}}. This prior embeds symmetric and topologically equivalent parts within the category into the same value (Geometric and Topological consistency). This implies that it profoundly understands the entire structure of given objects and common features of the category. Also, the topological prior is $SE(3)$ invariant, which is essential for robustly estimating the pose.

To obtain this potential prior, we enhance the surface embedding, an object-specific network introduced in SurfEmb~\cite{SurfEmb}. SurfEmb learns 2D-3D correspondences, allowing it to extract 2D embeddings that implicitly capture 3D geometric information. This network represents a 2D surface as a one-dimensional manifold, reducing the dimensionality caused by symmetry. Inspired by this approach, we hypothesize that dimension reduction while preserving the principal dimensions enables the extraction of embeddings that are invariant to shape variations within the same category.
To expand this object-specific network to a category-specific network, we retrain the network using synthesized category-level images. Specifically, we generated images from 3D models used in only the training set of the existing dataset~\cite{NOCS}. 
We establish this network as a backbone for obtaining topological prior and freeze both the training and inference stages in the main pipeline.

The backbone $f_s$ extracts topological embeddings from the input cropped object image $\mathbf{I}_{obj}$ and category vector $\mathbf{c}$. Given the complexity of object shapes at the category level, a deeper feature representation is required to accurately estimate poses. To achieve this, we apply a convolutional layer to further refine the output from $f_s$. This process enables the model to generate high-dimensional feature maps $\mathcal{F}^I_t \in \mathbb{R}^{H \times W \times \mathcal{C}}$ that preserve fine-grained details crucial for accurate category-level pose estimation. The operation is formulated as follows:
\begin{equation}
    \mathcal{F}^I_t = \textit{ConvLayer}(f_s(\mathbf{I}_{obj}, \mathbf{c}))
\end{equation}

Furthermore, the Topological prior captures detailed local features and provides a shared latent space that can effectively represent the global context, essential for addressing complex shape variations. To leverage the advantages of our prior, we use STViT \cite{STViT}, a well-suited vision transformer for capturing global context while preserving detailed local features. Thus, this module takes the topological embedding feature map $ \mathcal{F}^I_t$ as input and, following this, extracts the topological global feature $\mathcal{F}^g_t \in \mathbb{R}^{1 \times \mathcal{D}}$. This topological global context also mitigates the limitations of the 3D-GC network discussed in (\hyperref[subsec:HGF]{{\underline{\textcolor{blue}{Sec.} \textcolor{red}{3.3}}}}).

\subsection{Hybrid Graph Fusion} 
\label{subsec:HGF}

\begin{figure}[!t]
  \centering
  \includegraphics[width=\linewidth]{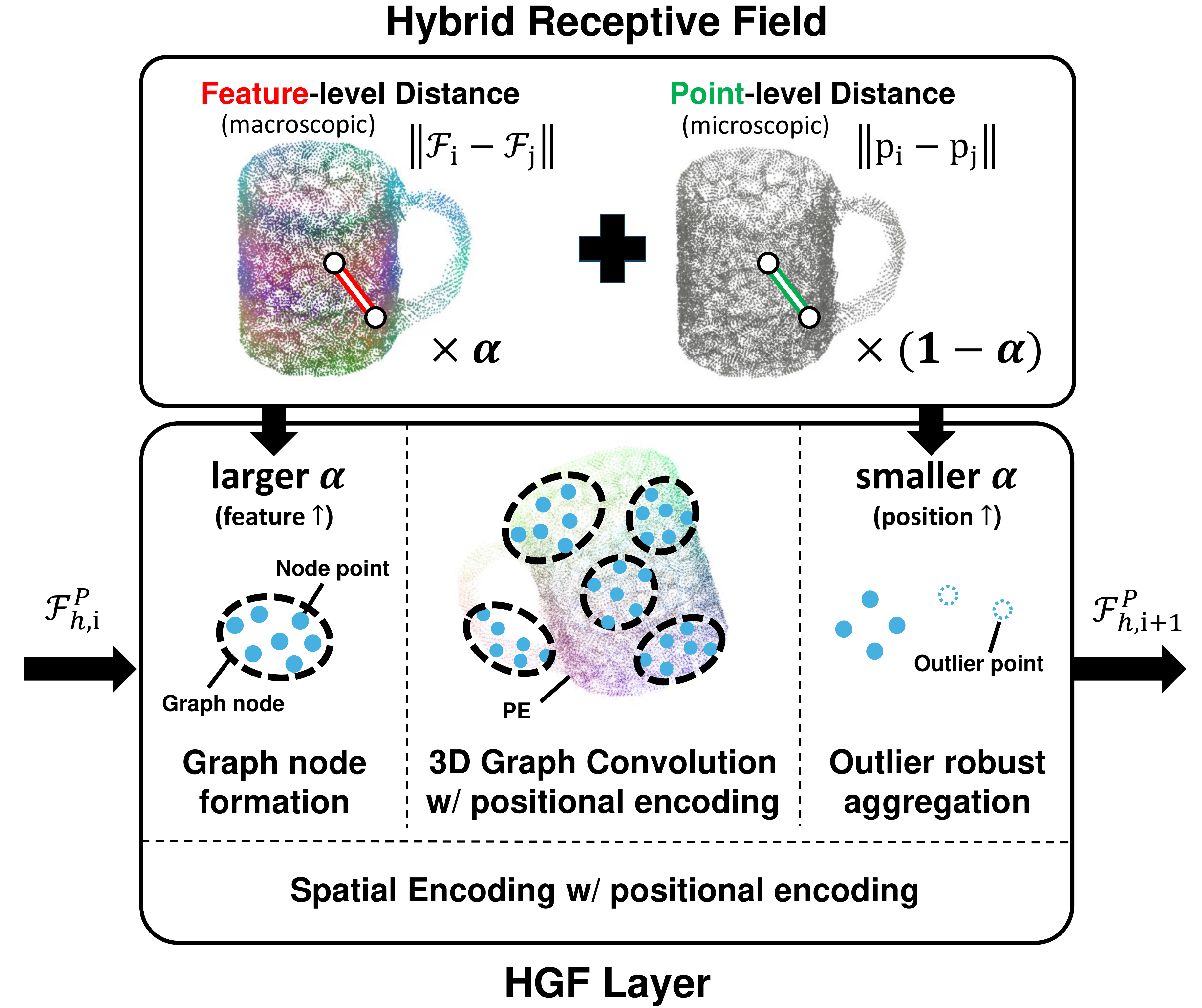}
  \caption{\textbf{Hybrid Graph Fusion Layer.} We define a hybrid receptive field that combines point-level and feature-level distances, enabling to percieve local-global geometry relationships. A two-path design enriches each path with positional encoding and 3D-GC, then applies outlier-robust aggregation. By adaptively adjusting the weight \(\alpha\) between point and feature distances, the HGF layer effectively handles scale/translation estimation, complex shapes, and partial occlusions, delivering refined pose-aware features.}
  \label{fig:hgf}
\end{figure}

\textbf{Hybrid Receptive Field.}
To capture local details and directly estimate 6D-pose without the matching procedure of shape prior methods, we adopt 3D graph convolution module~\cite{3D-GC}. Conventional 3D-GC \cite{3D-GC, FS-Net, GPV-Pose} methods integrate local receptive field in points cloud to build a graph node, based on Euclidean distance between points. In contrast, recent work \cite{HS-Pose} has shown that the receptive field in the feature space can capture more high-level relationships beyond distance.
We harmonize these two approaches by defining a hybrid receptive field (HRF) that combines both point-level and feature-level distances. Topological prior focus on macroscopic information; therefore, it lacks local geometric details such as surface distortion. The microscopic information from the point cloud can compensate for this limitation. Therefore, we use a weighting factor \(\alpha\) to balance their relative importance. The hybrid distance between two points \(p_i\) and \(p_j\) is defined as
\begin{equation}
\label{eq3}
D_{\alpha}(i,j) = \alpha\, D_{\text{feat}}(i,j) + (1-\alpha)\, D_{\text{point}}(i,j)
\end{equation}
where $D_{\text{feat}}(i,j)=\|\mathcal{F}_i-\mathcal{F}_j\|$ is the Euclidean distance between their corresponding feature representations and $D_{\text{point}}(i,j)=\|\mathbf{p}_i-\mathbf{p}_j\|$ is the Euclidean distance between the point clouds, as illustrated \hyperref[fig:hgf]{{\underline{\textcolor{blue}{Figure } \textcolor{red}{4}}}}. Before utilizing hybrid distance, we prepare the first feature map as follows. 

We first back-project the topological feature map  $\mathcal{F}^I_t \in \mathbb{R}^{H \times W \times D_1}$ to obtain  $\mathcal{F}^P_t \in \mathbb{R}^{N \times \mathcal{D}_1}$, which corresponds to the point cloud features. 
These are then concatenated with  $\mathcal{F}^P_{h,1} \in \mathbb{R}^{N \times \mathcal{D}_2}$, which is extracted from the 3D-GC using only on point distance ($\alpha=0$). The concatenated fusion feature $\mathcal{F}^P_f \in \mathbb{R}^{N \times ({\mathcal{D}_1+\mathcal{D}_2})}$ is used as an input of Fusion Stream.

\vspace{0.4cm}
\textbf{Fusion Stream.} Our 3D-GC based fusion module, termed \ac{HGF} module serves as capturing of local and global relationships, scale/translation awareness, and outlier-robust, which are the limitations of previous 3D-GC based methods. This module is composed of four HGF-layers and each HGF-layer follows a two-path design analogous to HS-layer \cite{HS-Pose}, as illustrated \hyperref[fig:hgf]{{\underline{\textcolor{blue}{Figure} \textcolor{red}{4}}}}.  
In the first path, we perform spatial encoding by applying positional encoding to the fused feature, providing essential spatial information for accurate size and translation estimation.
In the second path, the fused feature  $\mathcal{F}^P_{h,i}$ is passed through a hybrid receptive field that applies more weight to the feature distance by ${\alpha}_1$, forming graph nodes. These nodes are then enriched with positional encoding and fed into the 3D graph convolution layer. The output is processed outlier robust aggregation, where the hybrid receptive field with more emphasis on point distance by ${\alpha}_2$ is applied. This approach enhances robustness to outliers and captures a wider range of spatial geometric relationships, resulting in a more refined feature representation, denoted as \ $\mathcal{F}^P_{h,i+1}$.
Finally, the outputs of these two paths are combined via element-wise summation. Multiple HGF layers are stacked hierarchically, alternated with 3D max-pooling layers, to enhance both local detail and overall geometric consistency gradually. The final layer also outputs the global feature $\mathcal{F}^g_h$, which captures a larger geometric context. Ultimately, the features obtained from all layers are concatenated to form the final fused feature representation:
\begin{equation}
\label{eq4}
 \mathcal{F}^P_f = \mathcal{F}^P_{h,1} \oplus \mathcal{F}^P_{h,2} \oplus \dots \oplus \mathcal{F}^P_{h,5} \oplus \mathcal{F}^g_h.
\end{equation}

\subsection{Pose \& Size Estimator} \label{subsec:PSE}

Given the topological global-wise features $\mathcal{F}^g_t \in \mathbb{R}^{N \times \mathcal{D}_t}$ and the hybrid fused feature $\mathcal{F}^P_f \in \mathbb{R}^{N \times \mathcal{D}_f}$, we concatenate them point-wise and directly regress the rotation, as well as the residuals of size and translation. The process is formulated as follows:
\begin{equation}
\label{eq5}
\mathcal{F}_{pose} = \mathcal{F}^g_t \oplus \mathcal{F}^P_f
\end{equation}\begin{equation}
\label{eq6}
R, (t^*, S^*) = MLP_R(F_{pose}), MLP_S(F_{pose} \oplus P_{obj})  
\end{equation}
From \cite{HS-Pose}, we employ a decoupled rotation strategy for rotation estimation and estimate size and translation by predicting the residual size and translation between the ground truth and the mean position and size of the input point cloud. 

We use the extracted features during training to perform object pose regression, symmetric-based point cloud reconstruction, and bounding box voting from \cite{GPV-Pose}. Only the encoder and pose regression modules are employed during inference to predict the final pose and size parameters.

\section{Experiments}
\label{sec:experiments}

\subsection{Experimental Setup.}
\textbf{Datasets}
Following prior works~\cite{HS-Pose, AG-Pose, SecondPose}, we evaluate our approach on the NOCS benchmarks: REAL275 and CAMERA25~\cite{NOCS}, widely used for category-level 6D pose estimation. REAL275 is a challenging real-world dataset comprising 7K RGB-D images collected from 13 scenes with objects from 6 categories. It is split into a training set (4,300 images from 7 scenes) and a test set (2,750 images from 6 scenes), where each category includes three unseen instances. CAMERA25 is a large-scale synthetic dataset with 300K RGB-D images covering the same categories as REAL275. This dataset contains 1,085 object instances, with 25K images of 184 instances reserved for evaluation, while the remaining images are used for training.

\textbf{Implementation Details}
We adopt Mask R-CNN~\cite{MaskRcnn} for instance segmentation from our baseline HS-Pose~\cite{HS-Pose}, cropping each object-centric image to \(256\times256\) resolution and uniformly sampling 1,028 points from the corresponding point cloud. As a preliminary step, we retrain a category-level backbone using SurfEmb~\cite{SurfEmb} by rendering 450K synthetic images utilizing only CAMERA25's synthetic objects via Blender~\cite{Blender}, where 10 objects per scene are randomly placed to cover diverse intra-class variations. When training our framework on the NOCS dataset~\cite{NOCS}, we follow the strategy proposed in~\cite{NOCS}, using a mixture of 25\% real images from the REAL275 training set and 75\% synthetic images from the CAMERA25 training set.
For our model's hyperparameters, we select 15 neighbors and set \(\alpha_1 = 0.8\) and \(\alpha_2 = 0.2\) in our Hybrid Graph Fusion(HGF) module.
We maintain the same data augmentation strategy, loss terms, and training parameters as HS-Pose, and employ the Ranger optimizer~\cite{Ranger1, Ranger2, Ranger3} starting at a learning rate of \(1\times10^{-4}\). The learning rate is reduced via a cosine schedule during the final 28\% of training epochs. 
All experiments are conducted using a single NVIDIA Tesla H100 (80GB) GPU paired with two Intel Xeon Gold (32-core) processors.


\begin{table*}[htbp]
    \centering
    \caption{\textbf{Comparison with the state-of-the-art methods on the REAL275 dataset.} The best results are in bold, and the second-best results are underlined.}
    \label{tab:real275_comparison}
    \resizebox{1\textwidth}{!}{
    \begin{tabular}{l|c|c c|c c c c}
    \toprule
    \textbf{Method} & \textbf{Approach} & \textbf{$\text{IoU}_{50}$} & \textbf{$\text{IoU}_{75}$} & \textbf{5° 2 cm} & \textbf{5° 5 cm} & \textbf{10° 2 cm} & \textbf{10° 5 cm} \\
    \midrule
    SPD \cite{SPD}                     &  & 77.3 & 53.2 & 19.3 & 21.4 & 43.2 & 54.1 \\
    SGPA \cite{Sgpa}                &  & 80.1 & 61.9 & 35.9 & 39.6 & 61.3 & 70.7 \\
    CR-Net \cite{CR-Net}              &  & 79.3 & 55.9 & 27.8 & 34.3 & 47.2 & 60.8 \\
    SAR-Net  \cite{SAR-Net}       & Shape prior based & 79.3 & 62.4 & 31.6 & 42.3 & 50.4 & 68.3 \\
    SSP-Pose~\cite{Ssp-Pose}  &  &  82.3  & 66.3 & 34.7 & 44.6 & - & 77.8 \\
    RBP-Pose   \cite{RBP-Pose}  &  &  --  & 67.8 & 38.2 & 48.1 & 63.1 & 79.2 \\
    DPDN \cite{Self-DPDN}          &  & 83.4 & 76.0 & 46.0 & 50.7 & 70.4 & 78.4 \\
    \midrule
    FS-Net \cite{FS-Net}              &  & 81.1 & 63.5 & 19.9 & 33.9 & - & 69.1 \\    
    GPV-Pose \cite{GPV-Pose}              & 3D-GC based & 83.0 & 64.4 & 32.0 & 42.9 & 55.0 & 73.3 \\
    HS-Pose  \cite{HS-Pose}              &  & 82.8 & 75.3 & 46.2 & 56.1 & 68.9 & 84.1 \\
    \midrule
    AG-Pose \cite{AG-Pose}                & Keypoint based & \underline{83.7} & \underline{79.5} & 54.7 & 61.7 & \underline{74.7} & 83.1 \\
    SecondPose \cite{SecondPose}           & Semantic prior based & \textbf{83.8} & 77.7 & \underline{56.2} & \textbf{63.6} & \underline{74.7} & \underline{86.0} \\
    \midrule
    \textbf{THE-Pose (Ours)}          & Topological prior + 3D-GC & 83.4 & \textbf{79.6} & \textbf{56.7} & \underline{63.1} & \textbf{79.0} & \textbf{87.4} \\
    \bottomrule
    \end{tabular}}
\end{table*}

\textbf{Evaluation Metrics}
Following~\cite{NOCS, SPD}, we use the mean average precision (\textbf{mAP}) of the 3D Intersection over Union (IoU) with thresholds of 50\% and 75\% to evaluate the object's size and pose together. 
We evaluate the rotation and translation estimation performance using the metrics of {5\degree}, {10\degree}, 2 \textbf{cm}, and 5 \textbf{cm}, which means an estimation is considered correct if its corresponding error is lower than the threshold.
The pose estimation performance is also evaluated using the combination of rotation and translation thresholds: ({5\degree, 2 \textbf{ \cm}}), ({5\degree, 5 \textbf{ \cm}}), ({10\degree, 2 \textbf{ \cm}}), and ({10\degree, 5 \textbf{ \cm}}).


\subsection{Comparison with State-of-the-Art Methods}
\label{sec:comparison_sota}
\noindent \textbf{Results on REAL275 dataset.}
\hyperref[tab:real275_comparison]{\underline{\textcolor{blue}{Table}\textcolor{red}{1}}} compares our THE-Pose with recent state-of-the-art methods on the NOCS-REAL275 dataset. Our method consistently outperforms comparison methods across nearly all metrics, with highly competitive performance in the few remaining cases. First, we compare against the best method using shape prior DPDN~\cite{Self-DPDN}. Our method outperforms DPDN by a notable margin, particularly excelling at strict metrics: a +10.7\% gain in 5°\,2 \textbf{cm} and +12.4\% in 5°\,5  \textbf{cm}, demonstrating the strength of our topological prior over mean shape prior. Next, we evaluate performance relative to existing 3D-GC-based methods, highlighting HS-Pose~\cite{HS-Pose} as the best-performing baseline in this field. Compared to HS-Pose, our method improves the overall score by 35.8\%, with a +10.5\% lift on 5°\,2 \textbf{cm} and +4.3\% on IoU\(_{75}\). These notable gains confirm that our framework effectively addresses the scale and translation estimation limitations inherent in 3D-GC pipelines.
Finally, we compare our method against two state-of-the-art approaches. Our method outperforms HS-Pose and SecondPose across most metrics, notably exceeding HS-Pose by 35.1$\%$ and SecondPose by 7.2$\%$ overall. In particular, our topological prior demonstrates superior effectiveness compared to semantic priors from foundation models such as DINOv2~\cite{DINOv2}, compared with SecondPose.

\vspace{1mm}
\textbf{Results on CAMERA25 dataset  }
In \hyperref[tab:camera25_comparison]{{\underline{\textcolor{blue}  
{Table }\textcolor{red}{2}}}}, we present the quantitative results of THE-Pose on the CAMERA25 dataset. Our method achieves the highest accuracy under all evaluation metrics and notably outperforms the second-best method, AG-Pose~\cite{AG-Pose}. Specifically, our approach surpasses AG-Pose by +0.9\% on $\text{IoU}_{50}$, +1.1\% on $\text{IoU}_{75}$, +4.1\% on 5°\,2 \textbf{cm}, +4.3\% on 5°\,5 \textbf{cm}, +2.7\% on 10°\,2 \textbf{cm}, and +2.6\% on 10°\,5 \textbf{cm}. These improvements highlight the effectiveness of leveraging our topological prior to capturing both local and global geometry, ultimately leading to more robust performance on category-level 6D pose estimation.

\subsection{Improvements on 3D Graph Convolution by Topological Prior}
We demonstrate that our topological prior overcomes 3D-GC's limitations and boosts precise pose estimation in challenging scenarios.

\vspace{1mm}

\begin{figure*}[!t]
  \centering
  \includegraphics[width=0.9\textwidth]{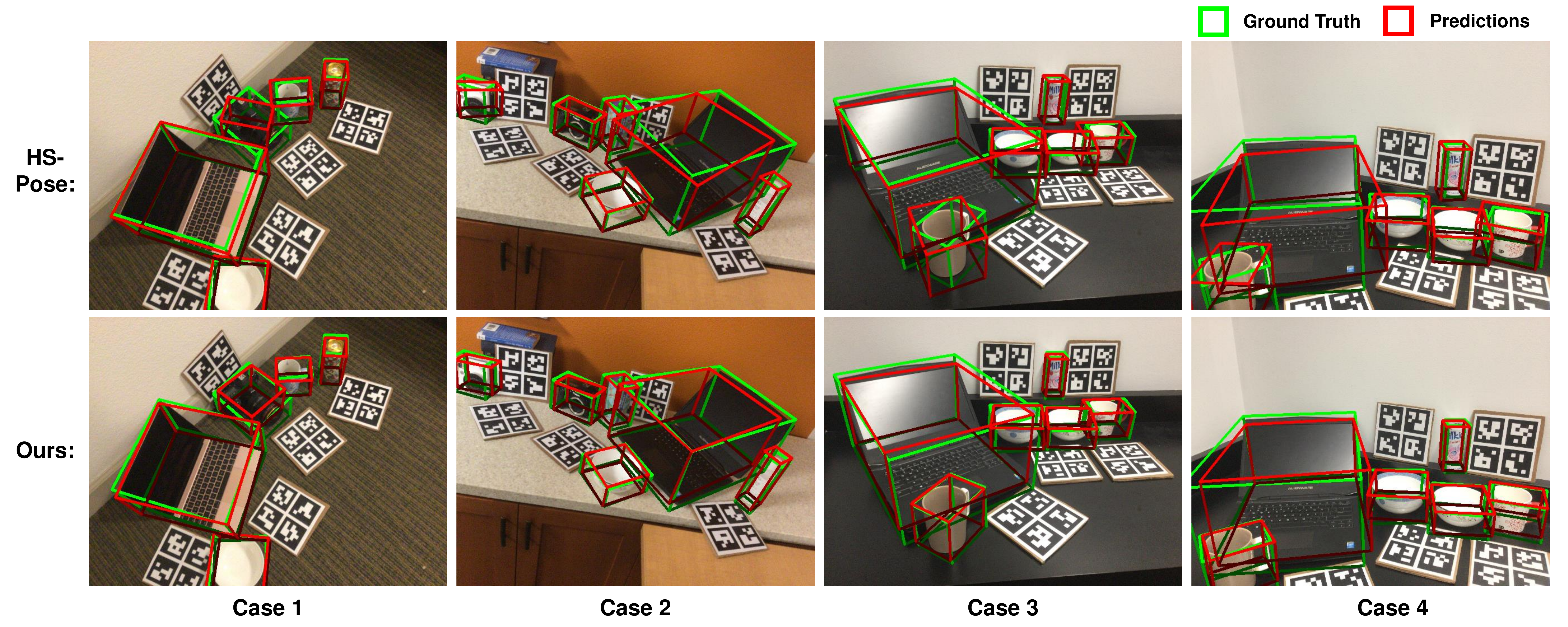}
  \caption{{\textbf{Qualitative comparison of our method and HS-Pose~\cite{HS-Pose} on the REAL275 dataset.} We compare the pose estimation predictions of our approach with the baseline method, HS-Pose~\cite{HS-Pose}. The ground truths are shown with green lines. Our method demonstrates notable improvements in accuracy, even in the presence of occlusions.}}

  \label{fig:Qulitative}
\end{figure*}

\noindent \textbf{Global context}
Our method achieves robust and accurate pose estimation even for complex objects and severe occlusions by leveraging topological consistency, which provides the global context of categories. In contrast, conventional 3D-GC methods mainly focus on local geometry, making it difficult to handle large shape variations or occlusion scenarios. 
As shown in case~1 and case~2 of \hyperref[fig:Qulitative]{\underline{\textcolor{blue}{Figure }\textcolor{red}{5}}}, the baseline approach~\cite{HS-Pose} fails to accurately estimate a camera’s 6D pose when faced with intricate structures and extensive intra-class variation. However, our approach remains robust under the same conditions. Moreover, case~4 illustrates that when certain areas of the mug are occluded, our method still infers the correct pose by leveraging global shape cues and maintaining consistent object representations. This robustness is also demonstrated by our ablation study on occlusion (\textbf{[AS-5]}). These results show how our topological prior, combined with local geometric information from point cloud features, effectively addresses the inherent limitations of 3D-GC-based methods in partial observation scenarios.

\textbf{Partial Symmetric Estimation }
Our method excels in scenarios where an object like a mug behaves asymmetrically (visible handle) yet becomes effectively cylindrical (occluded handle). In contrast, conventional 3D-GC approaches~\cite{HS-Pose} always treat the mug as fully asymmetric, leading to incorrect rotation estimates when the handle is obscured. As shown in case~3 of \hyperref[fig:Qulitative]{\underline{\textcolor{blue}{Figure }\textcolor{red}{5}}}, the baseline misaligns the bounding box under partial occlusion, whereas our topological prior leverages geometric consistency to distinguish truly symmetric surfaces from handle-induced asymmetry. Consequently, the network reduces the effective degrees of freedom in rotation once the handle disappears, focusing on the essential axis and delivering more accurate pose predictions.

\begin{table*}[t]
    \centering
    \caption{\textbf{Comparison with state-of-the-art methods on the CAMERA25 dataset.} The best results are in bold, and the second-best results are underlined.}
    \label{tab:camera25_comparison}
    \resizebox{0.8\textwidth}{!}{
    \begin{tabular}{l|c|c c|c c c c}
        \toprule
        \textbf{Method} & \textbf{Approach} & \textbf{IoU$_{50}$} & \textbf{IoU$_{75}$} & \textbf{5° 2 cm} & \textbf{5° 5 cm} & \textbf{10° 2 cm} & \textbf{10° 5 cm} \\
        \midrule
        SPD~\cite{SPD}               &  & 93.2 & 83.1 & 54.3 & 59.0 & 73.3 & 81.5 \\
        SGPA~\cite{Sgpa}             &  & 93.2 & 88.1 & 70.7 & 74.5 & 82.7 & 88.4 \\
        SAR-Net~\cite{SAR-Net}       & Shape prior based & 86.8 & 79.0 & 66.7 & 70.9 & 75.0 & 80.3 \\
        SSP-Pose~\cite{Ssp-Pose}     &  & --   & 86.8 & 64.7 & 75.5 & --   & 87.4 \\
        RBP-Pose~\cite{RBP-Pose}     &  & 93.1 & 89.0 & 73.5 & 79.6 & 82.1 & 89.5 \\
        \midrule
        GPV-Pose~\cite{GPV-Pose}      
            & \multirow{2}{*}{3D-GC based} 
            & 93.4 & 88.3 & 72.1 & 79.1 & --   & 89.0 \\
        HS-Pose~\cite{HS-Pose}
            & 
            & 93.9 & 89.4 & 73.3 & 80.5 & 80.4 & 89.4 \\
        \midrule
        AG-Pose~\cite{AG-Pose}       
            & Keypoint based
            & \underline{93.8} & \underline{91.3} & \underline{77.8} & \underline{82.8} & \underline{85.5} & \underline{91.6} \\
        \midrule
        \textbf{THE-Pose (Ours)}     
            & Topological prior + 3D-GC
            & \textbf{94.7} & \textbf{92.4} & \textbf{81.9} & \textbf{87.1} & \textbf{88.2} & \textbf{94.2} \\
        \bottomrule
    \end{tabular}}
\end{table*}

\begin{table*}[t]
    \centering
    \caption{\textbf{Ablation Study on REAL275 dataset.}}
    \renewcommand{\arraystretch}{1}
    \resizebox{0.8\textwidth}{!}{%
    \begin{tabular}{c|l|cc|cccc}
        \toprule
        \textbf{Row} & \textbf{Method} & \textbf{IoU$_{50}$} & \textbf{IoU$_{75}$} & \textbf{5° 2 cm} & \textbf{5° 5 cm} & \textbf{10° 2 cm} & \textbf{10° 5 cm} \\
        \midrule
        A0 & HS-Pose~\cite{HS-Pose} (baseline) & 82.1 & 74.7 & 46.5 & 55.2 & 68.6 & 82.7 \\
        \midrule
        B0 & A0 + TGC & 82.5 & 77.0 & 51.1 & 58.5 & 76.1 & 85.3 \\
        B1 & A0 + TGC + HGF & 83.3 & 78.0 & 52.9 & 59.3 & 76.8 & 86.6 \\
        \midrule
        C0 & B1 + P.E in spatial encoding & 83.7 & 80.0 & 54.2 & 60.0 & 77.5 & 87.3 \\
        \midrule
        D0 & C0 + P.E in graph nodes (\textbf{Full}) & 83.5 & 79.5 & 56.4 & 62.9 & 79.2 & 87.6 \\
        \midrule
        E0 & Neighbor number: 15 $\rightarrow$ 10 & 83.0 & 77.2 & 49.6 & 55.7 & 75.9 & 85.9 \\
        E1 & Neighbor number: 15 $\rightarrow$ 20 & 83.8 & 79.3 & 51.2 & 58.1 & 77.7 & 87.9 \\
        E2 & Neighbor number: 15 $\rightarrow$ 30 & 83.8 & 79.7 & 53.9 & 59.7 & 77.5 & 86.8 \\
        \midrule
        F0 & ${\alpha}_1$: 0.8  $\rightarrow$ 0.9 & 83.3 & 78.2 & 47.9 & 55.2 & 75.9 & 87.5 \\
        F1 & ${\alpha}_1$: 0.8  $\rightarrow$ 0.7 & 83.6 & 78.9 & 53.1 & 60.5 & 76.4 & 87.7 \\
        F2 & ${\alpha}_1$: 0.8  $\rightarrow$ 0.6 & 83.4 & 76.6 & 49.1 & 57.2 & 72.1 & 85.4 \\
        \midrule
        G0 & manual occlusion: 0.1 & 83.4 & 79.4 & 55.8 & 62.4 & 78.5 & 87.5 \\
        G1 & manual occlusion: 0.25 & 82.6 & 77.7 & 55.1 & 61.6 & 77.0 & 86.0 \\
        G2 & manual occlusion: 0.4 & 81.7 & 75.7 & 54.7 & 61.3 & 75.2 & 84.1 \\
        \bottomrule
    \end{tabular}}
    \vspace{1mm}
    \label{tab:ablation_study}
\end{table*}

\subsection{Ablation Studies}

To validate the effectiveness of our method, we perform several ablation studies on the REAL275 dataset~\cite{NOCS}. We incrementally integrate our proposed components (TGC, HGF, and positional encoding) into the baseline (HS-Pose) to evaluate their individual contributions. The complete ablation study results are presented in \hyperref[tab:ablation_study]{{\underline{\textcolor{blue}{Table }\textcolor{red}{3}}}}.

\noindent\textbf{[AS-1] Efficiency of using TGC and HGF modules.}
We demonstrate that integrating both the \textsc{TGC} and \textsc{HGF} modules into the 3D-GC framework (HS-Pose) significantly enhances pose estimation accuracy. As shown in \hyperref[tab:ablation_study]{{\underline{\textcolor{blue}{Table }\textcolor{red}{3}}}}, adding only \textsc{TGC} (Row [B0]) already improves strict metrics such as 5°,2, \textbf{cm} from 46.5 $\%$ [A0] to 50.1$\%$, highlighting the effectiveness of topological feature for capturing robust global context. Finally, combining both modules (Row [B1]) achieves a synergistic effect, delivering the best overall performance among all ablation settings. These results confirm that \textsc{TGC} effectively captures global context, while \textsc{HGF} enhances local and global geometric relationships, collectively addressing the limitations of baseline 3D-GC approaches.

\vspace{-0.5mm}
\noindent\textbf{[AS-2] Efficiency of positional encoding.}
Rows [C0] and [D0] in \hyperref[tab:ablation_study]{\underline{\textcolor{blue}{Table}\textcolor{red}{3}}} highlight the effect of integrating positional encoding (\textsc{PE}) in the HGF module. In row [C0], \textsc{PE} is applied to spatial encodings, while in [C1], it is incorporated into the graph node formation. Both configurations lead to notable improvements over row [B2], demonstrating that explicit positional information enhances precise geometric understanding. Finally, the combined approach in the row [D0], which applies \textsc{PE} to both spatial and node features, results in IoU$_{50}$ and IoU$_{75}$ of 83.5\% and 79.5\%, respectively, with a significant increase in strict metrics like 5°\,2\, \textbf{cm }(56.4\%). These results emphasize the effectiveness of positional encoding for accurate 6D pose estimation.

\noindent\textbf{[AS-3] Neighbor numbers.}
We study how the number of neighbors in our hybrid receptive field and outlier-robust aggregation module affects performance (rows [E0], [E1], [E2] in \hyperref[tab:ablation_study]{{{\underline{\textcolor{blue}
{Table }\textcolor{red}{3}}}}}). Reducing neighbors from 15 to 10 ([E0]) decreases all metrics, indicating that an overly local neighborhood misses the global context. In contrast, increasing to 20 or 30 neighbors ([E1], [E2]) generally improves both $\text{IoU}$ and precise metrics (5°\,2\, \textbf{cm}, 5°\,5\, \textbf{cm}) thanks to richer geometric cues. We note a slight trade-off: while 30 neighbors ([E2]) yields better $\text{IoU}_{75}$ and 5°\,2\, \textbf{cm} than 20 neighbors, its 10°\,5\, \textbf{cm} performance is marginally lower. This suggests that a moderate number can strike a favorable balance between global context and local detail, ensuring robust category-level pose estimation.

\noindent\textbf{[AS-4] Influence of \(\alpha\) in the hybrid receptive field.}
We investigate how varying the weighting factor ${\alpha}_1$ affects the hybrid receptive field in our HGF module (rows [F0-F2] in \hyperref[tab:ablation_study]{{\underline{\textcolor{blue}
{Table }\textcolor{red}{3}}}}. With a larger ${\alpha}_1$ [F0], the method emphasizes feature-based distances, capturing richer global context but sacrificing fine-grained local details. Consequently, metrics highlighting global alignment (10°\,2\, \textbf{cm}, 10°\,5\, \textbf{cm}) slightly improve, whereas stricter metrics (5°\,2\, \textbf{cm}) decrease. Conversely, a smaller ${\alpha}_1$ [F2] prioritizes point-based geometry, enhancing local detail at the expense of global consistency. An intermediate value (${\alpha}_1=0.7$ [F1], ${\alpha}_1=0.8$ [D0]) provides the optimal balance, demonstrating the importance of harmonizing local and global geometry for accurate category-level pose estimation.

\noindent\textbf{[AS-5] Robustness on occlusions.}
To further assess the robustness of our approach, we conduct the occlusion scenario by randomly masking a fixed percentage of pixels from both the object's cropped RGB and depth observations. Specifically, we evaluate our method under three different levels of occlusion: 10$\%$ ([G0]), 25$\%$ ([G1]), and 40$\%$ ([G2]) in \hyperref[tab:ablation_study]{{\underline{\textcolor{blue} {Table }\textcolor{red}{3}}}}. Under mild occlusion ([G0]), our method maintains stable performance with negligible degradation across all metrics. Even under severe occlusion ([G2]), the method preserves strong accuracy, with only modest decreases in strict metrics, as 5°,2, \textbf{cm}. This experiment highlights the effectiveness of our topological priors and the fusion of local geometric information from point cloud features, contributing significantly to robust pose estimation in challenging partial observation scenarios.

\section{Conclusion}
\label{sec:conclusion}

In this paper, we introduced \textbf{THE-Pose}, a novel category-level 6D object pose estimation framework that integrates a topological prior with 3D graph convolution (3D-GC). We define a topological prior that effectively bridges the gap between local geometry and global context, making it well-suited for both category-level pose estimation and 3D-GC. By efficiently fusing topological features with point cloud features, our approach overcomes the inherent limitations of 3D-GC methods, successfully handling large intra-class variations, complex object structures, and visual ambiguities. 
Extensive experiments on the REAL275 and CAMERA25 datasets demonstrate that THE-Pose significantly outperforms state-of-the-art methods, improving upon 3D-GC-based methods and surpassing previous best performances across key metrics. These results validate the effectiveness of combining topological priors with 3D-GC to achieve robust and accurate category-level pose estimation.

\newpage
{
    \small
    \bibliographystyle{ieeenat_fullname}
    \bibliography{main, rpm_packages/string-short}
}

\clearpage
\setcounter{page}{1}
\onecolumn
\begin{center}
    \Large
    \textbf{THE-Pose: Topological Prior with Hybrid Graph Fusion\\for Estimating Category-Level 6D Object Pose}    
    \vspace{0.4cm}
    \\
    Supplementary Material
\end{center}

\section{Visualization of Per-Category Topological Prior.}
In this supplementary material, we provide additional qualitative and quantitative results demonstrating the robustness of our proposed approach.

\begin{figure}[ht]
    \centering
    \includegraphics[width=\linewidth]{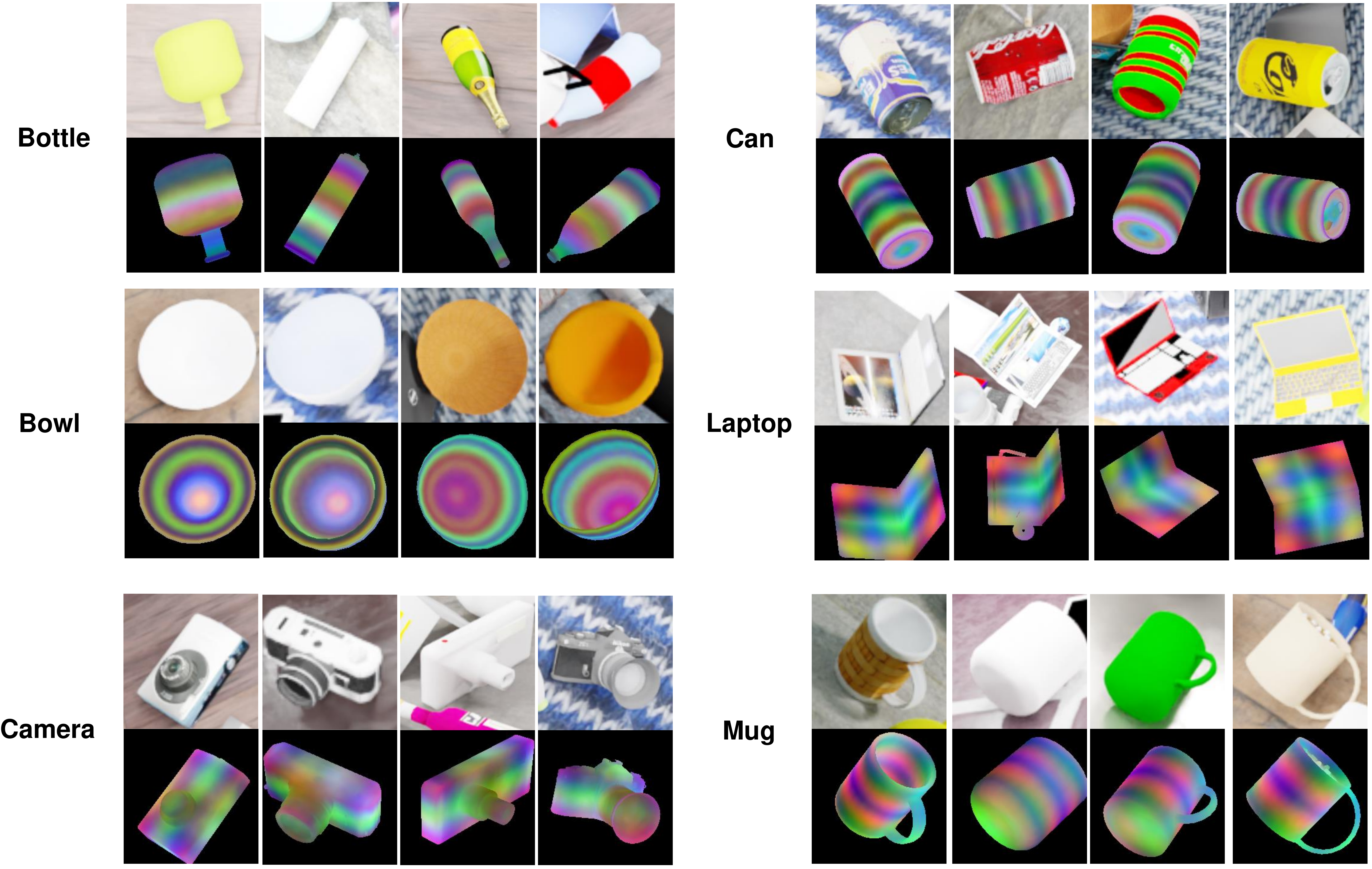}
    \caption{\textbf{Visualization of Per-Category Topological Prior.} }
    \label{fig:more1}
    \vspace{-2mm}
\end{figure}

In this section, we visualize the learned topological prior for each category and demonstrate that it effectively preserves consistent categorical information despite significant shape and color variations, as shown in \hyperref[fig:more1]{\underline{\textcolor{blue}{Figure }\textcolor{red}{6}}}. To further assess our prior's effectiveness on more complex and diverse object shapes, we conduct experiments on data synthesized from the CAMERA25 dataset~\cite{NOCS}.

\section{Additional Qualitative Results}
\label{sec:app_more}

\begin{figure}[ht]
    \centering
    \includegraphics[width=\linewidth]{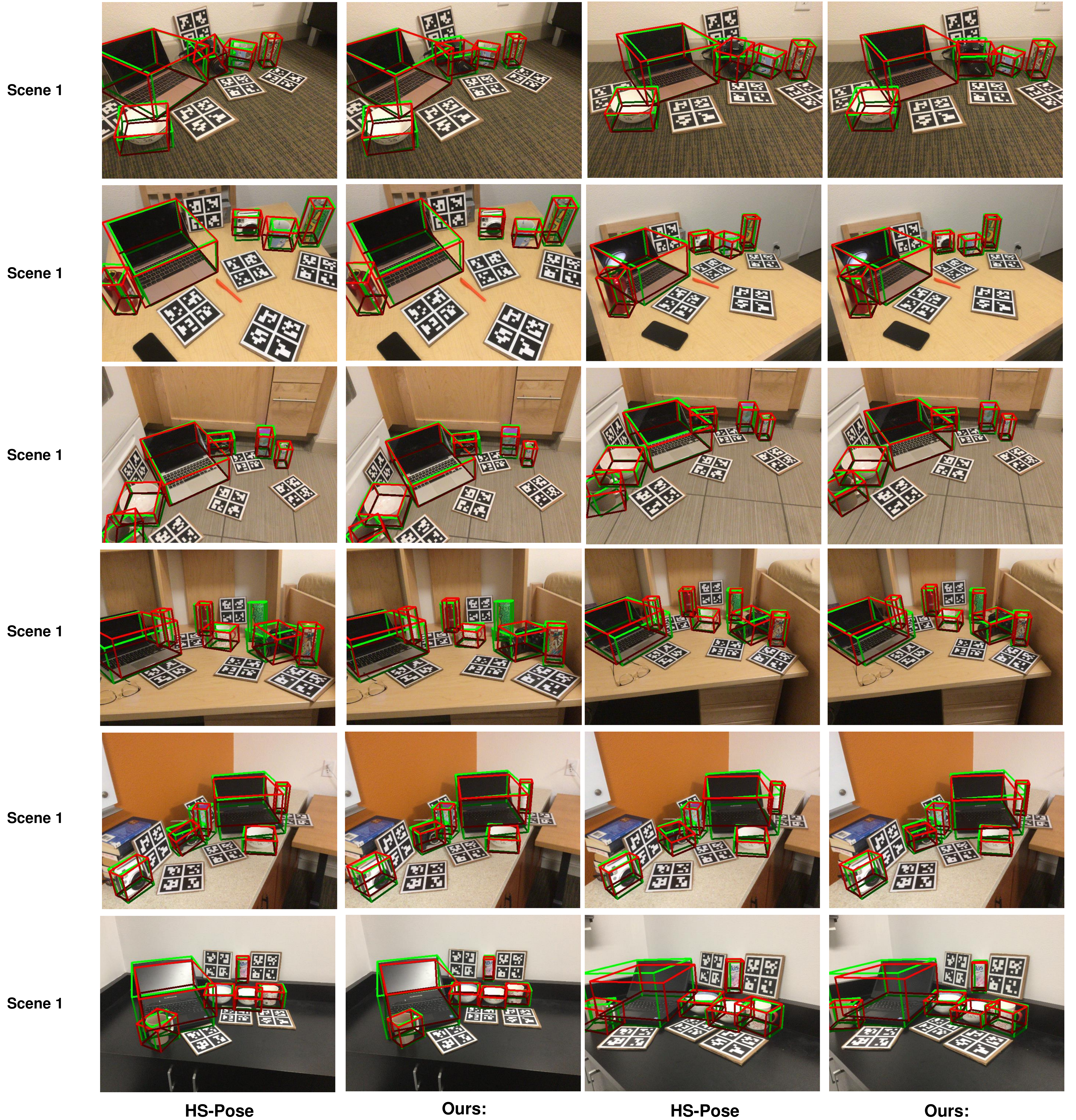}
    \caption{\textbf{More qualitative results of our method and the HS-Pose on the REAL275 dataset.} The predictions are illustrated with red lines and the ground truths are shown with green lines. Our method is more robust than the baseline in complex shapes and visual ambiguities.}
    \label{fig:more2}
    \vspace{-2mm}
\end{figure}

As shown in \hyperref[fig:more2]{\underline{\textcolor{blue}{Figure }\textcolor{red}{7}}}, our method consistently achieves stable and accurate pose estimation despite large intra-class variations. Notably, our framework remains robust under challenging scenarios, including complex-shaped objects (e.g., cameras), partial symmetry cases (e.g., mugs), and severe occlusions.

\section{Per-Category Results}
The detailed per-category results on the REAL275 and CAMERA25 datasets are illustrated in \hyperref[tab:real275_results_category]{{\underline{\textcolor{blue}  
{Table }\textcolor{red}{4}}}} and \hyperref[tab:camera25_results_category]{{\underline{\textcolor{blue}  
{Table }\textcolor{red}{5}}}}.

\begin{table*}[ht]
    \centering
    \caption{\textbf{Per-category results of our method on REAL275 dataset.}}
    \resizebox{1.0\textwidth}{!}{
    \begin{tabular}{l|c c c|c c c c c|c c| c c c}
        \toprule
        \textbf{category} & \textbf{IoU$_{25}$} & \textbf{IoU$_{50}$} & \textbf{IoU$_{75}$} & \textbf{5° 2cm} & \textbf{5° 5cm} & \textbf{10° 2cm} & \textbf{10° 5cm} & \textbf{10° 10cm} & \textbf{5°} & \textbf{10°} & \textbf{2cm} & \textbf{5cm} & \textbf{10cm} \\
        \midrule
        bottle & 57.7 & 57.7 & 53.8 & 62.3 & 68.8 & 87.2 & 94.8 & 99.6 & 85.8 & 100.0 & 87.2 & 94.8 & 99.6 \\
        bowl & 100.0 & 100.0 & 100.0 & 93.0 & 95.5 & 97.7 & 100.0 & 100.0 & 95.5 & 100.0 & 97.7 & 100.0 & 100.0 \\
        camera & 90.9 & 87.9 & 77.9 & 4.9 & 5.4 & 39.3 & 47.0 & 47.3 & 75.7 & 47.3 & 70.9 & 99.2 & 100.0 \\
        can & 71.4 & 71.4 & 71.2 & 77.1 & 79.3 & 97.0 & 98.7 & 98.7 & 82.3 & 99.8 & 97.1 & 98.8 & 98.8 \\
        laptop & 85.4 & 84.4 & 75.9 & 62.3 & 89.4 & 64.7 & 96.3 & 98.4 & 91.0 & 98.4 & 65.3 & 96.5 & 99.5 \\
        mug & 99.6 & 99.4 & 98.6 & 40.9 & 41.1 & 87.8 & 88.7 & 88.7 & 41.1 & 88.7 & 98.7 & 100.0 & 100.0 \\
        \midrule
        average & 84.2 & 83.5 & 79.6 & 56.7 & 63.1 & 79.0 & 87.6 & 88.8 & 66.6 & 89.0 & 86.2 & 98.2 & 99.7 \\
        \midrule
    \end{tabular}}
    \label{tab:real275_results_category}
\end{table*}

\begin{table*}[ht]
    \centering
    \caption{\textbf{Per-category results of our method on CAMERA25 dataset.}}
    \resizebox{1.0\textwidth}{!}{
    \begin{tabular}{l|c c c|c c c c c |c c |cc c c}
        \toprule
        \textbf{category} & \textbf{IoU$_{25}$} & \textbf{IoU$_{50}$} & \textbf{IoU$_{75}$} & \textbf{5° 2cm} & \textbf{5° 5cm} & \textbf{10° 2cm} & \textbf{10° 5cm} & \textbf{10° 10cm} & \textbf{5°} & \textbf{10°} & \textbf{2cm} & \textbf{5cm} & \textbf{10cm} \\
        \midrule
        bottle & 93.9 & 93.9 & 90.8 & 84.0 & 97.0 & 84.8 & 98.3 & 99.4 & 98.4 & 99.8 & 84.9 & 98.5 & 99.6 \\
        bowl & 96.9 & 96.8 & 96.7 & 99.5 & 99.6 & 99.7 & 99.8 & 99.8 & 99.7 & 99.8 & 99.7 & 99.9 & 99.9 \\
        camera & 95.0 & 93.2 & 88.2 & 70.3 & 75.6 & 81.8 & 90.9 & 91.0 & 75.7 & 91.0 & 88.0 & 99.5 & 99.7 \\
        can & 92.5 & 92.4 & 92.3 & 99.2 & 99.4 & 99.3 & 99.5 & 99.5 & 99.9 & 100.0 & 99.3 & 99.5 & 99.5 \\
        laptop & 98.4 & 97.7 & 92.9 & 78.1 & 90.2 & 81.2 & 94.5 & 97.9 & 93.6 & 98.0 & 82.7 & 96.4 & 99.9 \\
        mug & 94.1 & 93.9 & 93.3 & 60.3 & 60.5 & 82.1 & 82.3 & 82.3 & 61.5 & 82.9 & 99.1 & 99.5 & 99.6 \\
        \midrule
        average & 95.1 & 94.7 & 92.4 & 81.9 & 87.1 & 88.2 & 94.2 & 95.0 & 88.1 & 95.2 & 982.3 & 98.9 & 99.7 \\
        \midrule
    \end{tabular}}
    \label{tab:camera25_results_category}
\end{table*}


\end{document}